\documentclass[runningheads]{styling/llncs}

\usepackage[T1]{fontenc}
\usepackage{cite}
\usepackage{bm}
\usepackage{graphicx}
\usepackage{algorithm}
\usepackage[noend]{algpseudocode}
\usepackage{wrapfig}
\usepackage{appendix}
\usepackage{multirow}
\usepackage{booktabs}
\usepackage{tikz}
\usepackage{pgfplots}
\usepackage{caption}
\usepackage{subcaption}
\usepackage{wrapfig}
\usepackage{cleveref}
\usepackage{arydshln}
\pgfplotsset{compat=1.18}
\usepackage{url}

\definecolor{forestgreen(web)}{rgb}{0.13, 0.55, 0.13}
\definecolor{darkblue}{rgb}{0.0, 0.0, 0.55}

\begin{document}

\title{Interpretable Quantile Regression by Optimal Decision Trees}


\author{Valentin Lemaire\inst{1}\orcidID{0009-0007-2394-2526} \and 
Ga{\"e}l Aglin\inst{2}\orcidID{0000-0002-6760-4752} \and
Siegfried Nijssen\inst{2}\orcidID{0000-0003-2678-1266}
}

\authorrunning{V. Lemaire et al.}

\institute{Euranova, Rue Emile Francqui, Mont-Saint-Guibert, Belgium \and
Université Catholique de Louvain, Louvain-la-Neuve, Belgium}

\maketitle
\setcounter{footnote}{0} 
\begin{abstract}
    The field of machine learning is subject to an increasing interest in models that are not only accurate but also interpretable and robust, thus allowing their end users to understand and trust AI systems. This paper presents a novel method for learning a set of optimal quantile regression trees. The advantages of this method are that (1) it provides predictions about the complete conditional distribution of a target variable without prior assumptions on this distribution; (2) it provides predictions that are interpretable; (3) it learns a set of optimal quantile regression trees without compromising algorithmic efficiency compared to learning a single tree.

\end{abstract}

\section{Introduction}
    Recently, many studies have focused on rendering complex machine learning systems interpretable from a human perspective \cite{du2019techniques, Selvaraju_2019, fong2017interpretable}. Indeed, in many domains, the explanation is of equal interest as the accuracy of the prediction; especially in medical settings, business strategy settings, etc., both because one wants to understand why the prediction is made, but also to gain insights into the data by inspecting the model’s structure. However, most post-hoc techniques’ results remain complex to understand and often fail to explain the whole decision process of the AI system \cite{ieee_explainability_failings}. Therefore, work has been conducted in the direction of inherently interpretable models \cite{Letham_2015, rudin2019stop}. Within this family of inherently interpretable models falls the family of decision trees. Indeed, given a tree of reasonable depth, a human can quickly analyze it and see how and why the model predicts a particular class or value.
    
    Traditionally, decision trees are learned top-down by training them with a Mean Squared Error (MSE) heuristic, and by putting a single prediction in each leaf of the tree, corresponding to the mean of the target variable over the training examples. In this paper, we study {\em quantile regression}. A {\em quantile regression tree} does not predict a mean; for a given quantile parameter $q$, a quantile regression tree for that choice of $q$ would predict a value such that $q$ percent of the observed values are below the predicted value. Hence, for $q=10\%$ such a tree would underestimate, while for $q=90\%$ the tree would overestimate. This can be important in applications where well-motivated under- or overestimation is important, for instance, when using a tree to predict demand in retail, where the retailer would prefer to have a larger stock than the demand predicted by a standard regression tree.  However, there is no consensus on an efficient heuristic for quantile regression, which is why we interest ourselves in optimal decision trees, such as DL8 \cite{dl8}, DL8.5 \cite{dl8.5} and MurTree \cite{murtree}. Advantages of these methods include that (1) for trees constrained in depth these algorithms manage to find more accurate trees; (2) these algorithms can be used to learn trees without requiring the prior development of good heuristics for top-down tree induction.
    
    This latter characteristic makes optimal decision trees very suitable to quantile regression. However, an important weakness of learning a single quantile regression tree is that one would have to choose a single parameter $q$. Moreover, a single regression tree only provides limited insight in the complete conditional distribution of the target variable. In this paper we argue that it is often desirable to model the whole target distribution without making one choice for $q$ or without making prior assumptions with respect to the shape of this distribution: this increases trust in an AI system, provides insight in this target distribution, and allows to provide well-motivated under- or over-estimations of the target variable.  We propose to do so by learning a set of decision trees, each corresponding to a {\em quantile regression} tree. Indeed, predicting quantiles rather than the most likely value renders the models more robust to outliers \cite{q_robust_outliers} and doing this for many different quantiles gives information about the whole distribution. We introduce Quantile DL8.5 (QDL8.5). This method efficiently learns a {\em set} of optimal, shallow, and explainable decision trees for multiple quantiles, achieving high accuracy and interpretability. QDL8.5 addresses the complexity of choosing the right quantile in Quantile Regression by learning multiple trees for different quantiles. We show that this can be done with virtually no computational overhead compared to learning a single tree. The contributions of this paper are twofold: (i) We first propose an extension of the DL8.5 algorithm that enables it to perform quantile regression by one optimal tree per quantile while exploring the tree space only once, thus limiting drastically the time increase of learning many trees instead of one and (ii), we provide a robust assessment in terms of accuracy, execution time performance and interpretability.

\section{Related Work}
    Little work has gone towards performing quantile regression with interpretable decision trees. However, there are some notable works to mention. A very popular way to perform conditional quantile regression is Quantile Linear Regression (QLR) \cite{quantile_linreg}. Indeed, it is possible to derive a linear transform on features that finds the best linear fit to optimize quantile loss. While easily interpretable, this method is limited in terms of expressiveness. Quantile Regression Forests (QRF) \cite{qrf} represent a notable advancement in interpretable decision trees, sharing architecture with Random Forests \cite{breiman2001random}. Unlike traditional methods, QRF records samples associated with each leaf and predicts the empirical quantile based on the specified quantile during inference \cite{qrf}.  

    Cousins and Riondato introduced CaDET \cite{cadet}, a model that constructs interpretable decision trees or random forests predicting density functions within their leaves. They employ a selected statistical family (e.g., Gaussian, Pareto) instantiated in leaves and use empirical cross-entropy as the impurity measure for relevant splits. Despite fast training times due to heuristic-based tree growth, models like CaDET \cite{cadet} and QRF \cite{qrf} require more trees for sufficient expressiveness. Efficient heuristics exist for optimizing cross-entropy or mean squared error, but consensus is lacking for heuristics optimizing quantile loss in decision tree growth. Some other works have used more expressive but less interpretable methods to estimate conditional quantiles. Wang et al. \cite{wang2019} propose to use different standard regression models (random forests, support vector regressor and gradient-boosted decision trees) and combine them linearly to optimize quantile loss. Finally, they use Kernel Density Estimators (KDE) on the quantiles to predict a conditional probability density function. Similarly, Zhang et al. \cite{zhang2020} combined linearly different methods that already perform quantile regression including QLR and QRF. They also combine those quantile results into pdfs using KDE. Both of these methods combine models that are complicated to interpret, making them uninterpretable.

    Another approach might be to use model trees like those introduced by Quinlan et al. \cite{quinlan}, where, rather than having a single prediction in the leaf, a simple model (e.g. QLR) is fitted on the mapped samples. While this is purposeful for point predictions, it increases complexity of the models and it does not help to give a fuller picture of a sample's distribution without prior assumptions on said distribution. These will therefore not be included in the experimental evaluation.

\section{Background}
    As a first step into technical background, let us introduce quantiles.
    \begin{definition}
        Given a probability density function (pdf) $f$ describing a distribution, from which can be derived a cumulative density function (cdf) $F$, and given a quantile $q$, the corresponding quantile value $y_q$ is such that the probability that a realization $y \in \mathcal{Y}$ of the random variable $Y$ is lower than $y_q$ is $q$.
    \end{definition}

    Empirically, given a set of realizations $\bm{y}$, it was shown \cite{qrf} that the quantile value for a given quantile $q$ is the value $\hat{y}_q$ that minimizes the quantile loss:

    \begin{equation}\label{eq: quantile loss def}
        QL_q(\bm{y}, \hat{y}_q) = \sum_{i = 1}^{|\bm{y}|} \max \left\lbrace q (\hat{y}_q - y_i), (1 -q)  (y_i - \hat{y}_q) \right\rbrace.
    \end{equation}

    Quantile Regression is identical to standard Regression but the loss to optimize is the loss in Eq. \ref{eq: quantile loss def}. Another main building block of this work is optimal decision trees, and more specifically, the DL8 and DL8.5 implementations of these. DL8 \cite{dl8} and its newer, more efficient version DL8.5 \cite{dl8.5} are algorithms that enable the learning of the best tree that optimizes any additive loss function under constraints of maximum depth, minimum support, etc. DL8.5 works as follows. Given a binary featured dataset, it goes through each feature following a dynamic programming principle in a branch-and-bound manner.


    In these models, data is represented as itemsets, i.e. a collection of positive $f$ or negative $\neg f$ items for each feature $f$. When a decision tree is built, it splits each feature into its positive and negative branches, thus building larger itemsets as it goes down in the trees. For the remainder of this paper, we will consider our datasets to be binary as any dataset can be binarized (categorical features turned into one-hot encoded features and continuous features turned into binary bins). Each leaf of a decision tree will map to a certain subset of the samples that all contain the corresponding itemset. For example, if a leaf contains the itemset $\{ a, \neg d, g \}$, all samples having ones for features $a$ and $g$ and a zero for feature $d$ will be mapped to that itemset (and leaf). 
    
     Algorithm \ref{algo:dl8.5} shows how DL8.5 works. At each level of the search in DL8.5, a split is performed on each feature (line \ref{line:for attribute split}) and the overall error associated with each split is computed, going down recursively in the tree, always keeping track of the path that leads to the lowest error. This is an exhaustive search, but the search space is pruned by efficient lower and upper bounding. Indeed, the best errors of the branches at the same level act as upper bounds for the following negative branches (line \ref{line:left_ub}). The upper bound used for the positive branches that are explored after the negative ones is the difference between the previous upper bound and the error of the negative branch, as errors are additive (line \ref{line:postiveub}). To summarize in a sentence: \textit{at each step of the exploration, a new feature is chosen to be added to the itemset, the corresponding error is found for the two branches (positive and negative) and this feature is saved if it yields the best error so far. This is done by pruning as many unnecessary branches as possible.}
    \begin{algorithm}[!htbp]
        \caption{DL8.5($maxdepth$, $minsup$)}
        \scriptsize
        \label{algo:dl8.5}
        \begin{algorithmic}[1]
            \State \textbf{struct} $BestTree\{lb: float, tree:Tree, error:float\}$
            \State $cache \gets HashMap<Itemset, BestTree>$
            \State $best\_solution \gets DL8-Recurse(\emptyset, +\infty, 0)$
            \State \textbf{return} $best\_solution.tree$
            \Procedure{$DL8.5-Recurse$}{$I$, $ub$}
               \If{$leaf\_error(I) = 0$ or $|I| = maxdepth$ or time-out is reached}
                    \State \textbf{return} $BestTree(ub, make\_leaf(I), leaf\_error(I))$
               \EndIf
               
               \State $solution \gets cache.get(sort(I))$\label{line: cache get}
               
               \If{$solution$ was found}
                    \If {solution.tree $\neq$ \texttt{NO\_TREE} or $ub \leq solution.lb$}
                        \State \textbf{return} $solution$ \label{line:cache solution}
                    \EndIf
               \EndIf
               
               \State $(\tau, b, left\_ub) \gets (\texttt{NO\_TREE}, +\infty, ub)$
               
               \For{all attributes $i$ in a well-chosen order}\label{line:for attribute split}
                    \If{$cover(I \cup \{i\}) \geq minsup$ and $cover(I \cup \{\neg i\}) \geq minsup$}
                        \State $sol_1 \gets DL8.5-Recurse(I \cup \{\neg i\}, ub)$
                        \If {$sol_1.tree = \texttt{NO\_TREE}$}
                            \State \textbf{continue}
                        \EndIf
                        
                        \If {$sol_1.error < left\_ub$}\label{line:left_ub}
                            \State $sol_2 \gets DL8.5-Recurse(I \cup \{i\}, left\_ub - sol_1.error)$\label{line:postiveub}
                            \If {$sol_2.tree = \texttt{NO\_TREE}$}
                                \State \textbf{continue}
                            \EndIf
                            
                            \State $feature\_error \gets sol_1.error + sol_2.error$
                            
                            \If {$feature\_error < left\_ub$}
                                \State $\tau \gets make\_tree(i, sol_1.tree, sol_2.tree)$
                                \State $b \gets feature\_error$
                                \State $left\_ub \gets b$
                            \EndIf
                            \If {$feature\_error = solution.lb$}
                                \State \textbf{break}
                            \EndIf
                        \EndIf 
                    \EndIf
               
               \EndFor
               \State $solution \gets BestTree(ub, \tau, b)$
               \State $cache.store(I, solution)$\label{line: cache store}
               \State \textbf{return} $solution$
            \EndProcedure
        \end{algorithmic}
    \end{algorithm}
    DL8.5 also makes use of a cache (line \ref{line: cache get} and \ref{line: cache store}). Indeed, each node in the search tree can be seen as an itemset of positive or negative features that must be present in the samples mapped to that node. However, an itemset is an unordered collection of positive or negative features, meaning that it is possible to encounter the same itemset in different parts of the search tree. Indeed, exploring $a$ after $\neg b$ yields the same itemset as exploring $\neg b$ after $a$. The cache allows us to save the best trees for these itemsets and therefore avoid performing the same computation twice. If, for a particular itemset, the search yielded no result, this is also saved with the upper bound that was used, as a later search might ask for the same itemset with an equal or lower upper bound. This search can be avoided as it is already known it will give no result. 

    Demirovi\'{c} et al. \cite{murtree} proposed some optimisations to further the pruning by a better lower bounding and using a special computation for depth-two trees. Some of these optimisations are transferable to regression and have been included in the implementation but not shown in the algorithm as they are not principal for this paper.  This dynamic programming approach combined with efficient branching and bounding allows the algorithm to ensure finding the best decision tree under the given constraints. This formulation of the DL8.5 algorithm is already able to find the optimal decision tree optimizing quantile loss as it is an additive function. The only change that has to be brought is the implementation of the quantile loss as the leaf error function.

\section{Quantile DL8.5}
    In this section, we will show how to extend the DL8.5 learning algorithm to perform \textit{simultaneous} quantile regression\footnote{Source code is available at \url{https://github.com/valentinlemaire/pydl8.5}}. This change in the algorithm will allow it to learn, for each given quantile, an optimal decision tree while only exploring the search space once, thus utilizing for all quantiles the common parts of the search trees. This will be specially marked if the obtained decision trees are very similar as they will have resulted from similar searches among possible decision trees. The output of the algorithm is, for each sample, an array of quantile values, each corresponding to a different quantile. Given enough quantiles, they describe most of the conditional distribution of a sample. The more quantiles, the more this full distribution will be precise\footnote{We however recommend setting the number of quantiles to be lower than the minimum support in each leaf to avoid skewed estimations}.

    \subsection{Simultaneous Tree Learning}
        The quantile loss in equation (\ref{eq: quantile loss def}) is parametric, meaning that its value is dependent on the corresponding quantile. This also means that a tree that is optimal for one quantile is not guaranteed to be optimal for another. For this reason, we need to learn an optimal tree for each quantile. However, it is reasonable to assume that these trees will not be very different from each other, especially for close quantiles as they will describe close parts of the conditional distribution. Therefore, running independent searches for the optimal trees would be inefficient as the searches will go through many common itemsets for the different quantiles. For this reason, we have changed the DL8.5 algorithm to enable it to learn the optimal decision tree for different quantiles while only exploring each itemset once, no matter how many trees have to be learnt. 
    
        To enable DL8.5 to learn many trees at once a few changes had to be brought to the algorithm. Algorithm \ref{algo:qdl8.5} shows these changes. This algorithm is essentially the same as Algorithm \ref{algo:dl8.5} except that all conditions of pruning change and some computations need to be performed once for each quantile. Indeed, in QDL8.5, a branch can only be pruned if all the searches relating to all the different quantiles do not give any results. Thus, line \ref{line: can return} shows a call to a $can\_return$ function. This function returns true if, \textit{for all quantiles}, there is either no solution, either the upper bound is lower than the lower bound or the leaf error has attained its lower bound.

        During the search, when considering an itemset, for each quantile we consider each possible feature. However, if we consider a particular itemset for a particular quantile, we compute the quantile values and quantile loss for all the quantiles. Therefore, if we get back to that itemset for a different quantile later in the search, the quantile value and loss will already be saved to the cache and therefore no additional computation will be needed. Section \ref{sec:o(n) for quantile comp} shows how we can compute many quantile values and losses without additional computational cost. It can also be seen that for each quantile and for each itemset explored, there are lower and upper bounds and associated errors.

        \begin{algorithm}[!hbtp]
        \scriptsize
        \caption{QDL8.5($maxdepth$, $minsup$)}
        \label{algo:qdl8.5}
        \begin{algorithmic}[1]
            \State \textbf{struct} $BestTree\{\bm{lbs}: vector<float>, \bm{trees}:vector<Tree>, \bm{errors}:vector<float>\}$
            \State $cache \gets HashMap<Itemset, BestTree>$
            \State $best\_solution \gets QDL8.5-Recurse(\emptyset, +\infty^{|\bm{q}|})$
            \State \textbf{return} $best\_solution.tree$
            \Procedure{$QDL8.5-Recurse$}{$I$, $\bm{ubs}$}
               \State $solution \gets cache.insertOrGet(sort(I))$
               
               \State $\bm{leaf\_errors} \gets quantile\_errors(I, \bm{q})$
               \If{$|I| = maxdepth$ or time-out is reached} 
                    \State \textbf{return} $BestTree(solution.\bm{lbs}, make\_leafs(I), \bm{leaf\_errors})$
               \EndIf 
               
               \If {$can\_return(solution, \bm{ubs}, \bm{leaf\_errors})$}\label{line: can return}
                    \State \textbf{return} $solution$
                \EndIf
               
               \For{all attributes $F$ in a well-chosen order} 
                    \If{$cover(I \cup \{f\}) \geq minsup$ and $cover(I \cup \{\neg f\}) \geq minsup$}
                        \State $sol_1 \gets QDL8.5-Recurse(I \cup \{\neg i\}, \bm{ubs})$
                                                
                        \If{all $sol_1.\bm{trees}$ are \texttt{NO\_TREE}} 
                            \textbf{continue} \label{line: qdl85 sol1 no tree}
                        \EndIf
                        
                        \If {$\exists i \in \{1, 2, \dots, |\bm{q}|\}: sol_1.errors_i < solution.errors_i$}\label{line: qld85 if exists i...}
                            \State $sol_2 \gets QDL8.5-Recurse(I \cup \{i\}, \bm{ubs} - \bm{sol_1.errors})$
                                                        
                            \If {all $sol_2.\bm{trees}$ are \texttt{NO\_TREE}}\label{line: qdl85 sol2 no tree}
                                \textbf{continue}
                            \EndIf

                            \For{$i \in \{1, 2, \dots, |\bm{q}|\}$}\label{line: qdl85 for i in q 3}
                                \State $feature\_errors_i \gets sol_1.errors_i + sol_2.errors_i$
                                
                                \If{$feature\_errors_i < solution.errors_i$}
                                    \State $solution.trees_i \gets build\_tree(F, sol_1.trees_i, sol_2.trees_i)$
                                    \State $solution.errors_i \gets feature\_errors_i$
                                    \State $ubs_i \gets feature\_errors_i$
                                \EndIf
                            \EndFor
                            
                            \If {all $\bm{feature\_errors} = solution.\bm{lbs}$}\label{line: qdl85 all optimal for feature}
                                \textbf{break}
                            \EndIf
                            
                        \EndIf 
                    \EndIf
               \EndFor
               
               \For{$i \in \{1, 2, \dots, |\bm{q}|\}$}\label{line: qdl85 for i in q 4}
                    \State $solution.lbs_i \gets ubs_i$
               \EndFor
               \State \textbf{return} $solution$
            \EndProcedure
        \end{algorithmic}
    \end{algorithm}

    \subsection{Efficient Quantile Loss Computation}\label{sec:o(n) for quantile comp}
        The most costly operation in the DL8.5 (and QDL8.5) method is traversing the data to compute the predictions and the errors for each node (itemset) explored. It is therefore primordial to make that operation as efficient as possible. The naive way would be to apply equation (\ref{eq: quantile loss def}) which can be computed in $\mathcal{O}(N)$ time for each quantile, leading to a calculation time of $\mathcal{O}(|\bm{q}|N)$, where $N$ is the number of samples mapped to that itemset, i.e. the cover of that itemset, and $|\bm{q}|$ is the number of quantiles.  As a first step towards computational efficiency, the complete data can be sorted according to the $\bm{y}$ values before starting the tree search. Indeed, the empirical estimation of a quantile is defined as

        \begin{equation}\label{eq: numpy quantile estimation}
            \hat{y}_q = y_{\lfloor h \rfloor} + (h - \lfloor h \rfloor) (y_{\lceil h \rceil} - y_{\lfloor h \rfloor})
        \end{equation}

        with $h = q(N-1) + 1$. This can be computed in $\mathcal{O}(1)$ time if the array of values $\bm{y}$ is sorted. This is also true for any subset of values as a subset of a sorted array is itself sorted. 
        Using this sorted array we can compute the quantile values corresponding to all the quantiles $\bm{q}$ in $\mathcal{O}(|\bm{q}|)$ time.
        We can also notice that the quantile loss formulated in equation (\ref{eq: quantile loss def}) can be rewritten as
        \begin{equation} \label{eq: quantile loss rewrite}
            QL_q(\hat{y}_q, \bm{y}) = (1 - q)\sum_{i: y_i \leq \hat{y}_q}y_i - (1-q)\sum_{i: y_i \leq \hat{y}_q} \hat{y}_q + q\sum_{i: y_i > \hat{y}_q} \hat{y}_q - q\sum_{i: y_i > \hat{y}_q}y_i
        \end{equation}
        Notice there is no index on $\hat{y}_q$ as the prediction is the same for all samples mapped to a leaf. Using this formulation we can see that the only need for the data is to store, for each quantile, the sum of the $y$ values that are above the prediction and the sum of those under the prediction. This can be used to compute the errors for all quantiles by only traversing the data once. Indeed, it is possible to go through the samples in increasing order (since the data is sorted) and bin the different samples according to the quantile values. Then, with a loop over the quantiles, it is possible to compute the sum of elements above and below each quantile value, thus allowing us to compute the quantile loss for each quantile efficiently. Using this technique, we get a temporal complexity of $\mathcal{O}(N + |\bm{q}|)$. However, it makes little sense to compute more quantiles than there are samples, so we always recommend setting $|\bm{q}| \leq minsup$, with $minsup \leq N$, thus making the temporal complexity of this $\mathcal{O}(N)$.
    
    \textit{Probability Density Estimates}
        With this implementation of the algorithm, the output at prediction time is an array of conditional quantile values, each corresponding to a different quantile. Given enough quantiles, they describe most of the conditional distribution of a sample. For a more visual interpretation, we can combine those quantiles in a distribution function using Kernel Density Estimators in the same way as in other works \cite{wang2019, zhang2020}. This adds two new parameters, the kernel and the width of these kernels. In the remainder of this paper, we'll use Scott's rule for kernel widths, which was shown to be optimal when the underlying distribution is Gaussian \cite{scotts_rule}. Even though we cannot make this assumption, we will use this method to estimate kernel widths. Correspondingly, we'll use Gaussian kernels.

\section{Experiments}
    For our experiments, we consider three different aspects of the performance of our model: (i) the accuracy by measuring how well the outputs of the model (estimated quantiles and pdfs) describe the actual data, (ii) the efficiency in terms of execution time and (iii) a study of how interpretable the obtained models are. We consider our main competitors to be CaDET \cite{cadet} and Quantile Random Forests \cite{qrf} as they are both ensemble methods using decision trees and outputting distribution information. For our experiments, we have chosen 4 datasets; one synthetic and 3 real-world. The synthetic dataset is generated as follows: using 9 binary features we created 15 categories represented by a combination of those features. Each category has its associated Gaussian distribution, each with a different mean and standard deviation, from which target values were drawn. This dataset was created to have a dataset with a known conditional probability distribution. In addition to it, we measured quality on three real-world datasets: Air Quality, Solar Flares, and Stock Portfolio Performance as they are widely used benchmark datasets of varying sizes, with low dimensionality that have categorical features (thus losing less information in binarization). All our experiments were performed on a machine running Intel(R) Xeon(R) Gold 6134 CPU@3.20GHz processor with 32 physical cores and 128GB RAM with Ubuntu 18.04.6 operating system. 
    
    \subsection{Metrics}
        To ensure the quality of the predictions we use different metrics. First is Mean Integrated Squared Error (MISE), which measures the integral of the squared difference between the true and predicted distributions. This can only be done for the synthetic dataset of which we know the true distribution. We also use Mean Quantile Error (MQE), which is the mean of all the different quantile errors over the different quantiles. We also measured the Negative Log Likelihood (NLL) of the samples with respect to the predicted distribution function and finally, the Continuous Ranked Probability Score (CRPS), which measures the integral of the squared difference between the predicted cdf for a sample and the step function for the actual realization of that sample.

        Regarding interpretability, we wish to demonstrate that while we learn many trees which can impede on overall interpretability, the resulting trees are mostly similar and analyzing just a few trees would give sufficient insights to the analyst. To this effect, we evaluated the partitions of the training dataset generated by each tree and measured a Jaccard index on all pairs of these partitions. With this metric, we would expect to see block matrices attesting that trees corresponding to close quantiles are indeed similar, thus attesting to overall interpretability.
        
    \subsection{Results}
        This section will illustrate our experiments and related observations. 
        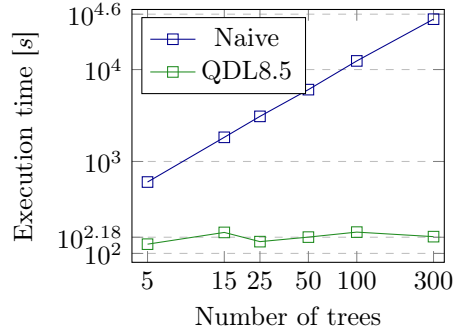
\begin{wrapfigure}{R}{0.5\textwidth}
            \centering
            \begin{tikzpicture}
            \begin{axis}[
                width=0.47\textwidth,
                xlabel={Number of trees},
                ylabel={Execution time [$s$]},
                xmin=4, xmax=350,
                ymin=80, ymax=45000,
                xtick={5,15,25,50,100,300},
                ytick={100,150,1000,10000,40000},
                ymode=log,
                xmode=log,
                log basis x=2,
                log basis y=10,
                xticklabel=\pgfmathparse{2^\tick}\pgfmathprintnumber{\pgfmathresult},
                legend pos=north west,
                ymajorgrids=true,
                grid style=dashed,
            ]
            
            \addplot[
                color=darkblue,
                mark=square,
                ]
                coordinates {
                (5.0,597.77)(15.0,1831.6599999999999)(25.0,3095.14)(50.0,6046.333333333333)(100.0,12408.28)(300.0,35297.62)
                };
    
            \addplot[
                color=forestgreen(web),
                mark=square,
                ]
                coordinates {
                (5.0,126.06)(15.0,169.22)(25.0,134.32)(50.0,150.47)(100.0,170.54999999999998)(300.0,152.22)
                };
            \legend{Naive,QDL8.5};
                
            \end{axis}
            \end{tikzpicture}
            \caption{Running times for naive and efficient versions of QDL8.5. Both axes are in \textit{log} scale.}
            \label{fig:efficiencyplot}
        \end{wrapfigure}
        \textit{Quality} Our first experiment concerns the quality of the obtained regressors. In table \ref{tab:quality_metrics}, we show, for each dataset and each model, the different quality metrics. All three methods get good results. QDL8.5 performs either as the best model or second best model on all metrics and datasets, often being close to the best result when not achieving it. From these observations, we can conclude that QDL8.5 is competitive with existing methods and matches the state of the art in decision trees that predict (some form of) conditional distributions. 
        \begin{table}[!h]
            \centering
            \caption{Quality metrics on all 4 datasets. For all metrics, lower is better. \textbf{Bold} means best, \underline{underlined} means second best. We performed hyperparameter tuning with Bayesian search (20 trials) for each dataset and method.}
            \begin{tabular}{ll c|c ccccc}
                \toprule
                Datasets & Metrics & & \multicolumn{6}{c}{Models}\\
                 & &\;\; &\; & Quantile RF &\; & CaDET RF & \; & Quantile DL8.5 \\
                \midrule
                \multirow{5}{*}{Synthetic dataset} & MISE & & & \underline{0.121} & & 0.122 & &  \textbf{0.120} \\
                 & NLL & & & \underline{2.16} &  & \textbf{2.15} &  & \textbf{2.15} \\
                 & MQE & & & 0.0700 &  & \underline{0.0616} & & \textbf{0.0603} \\
                 & CRPS & & & \textbf{1.22}  & & 1.25 &  & \underline{1.24} \\
                 \cdashline{2-9}
                 & \textit{n. trees / depth} & & & \textit{100 / 5}  & & \textit{50 / 5} & & \textit{100 / 4}\\
                 \midrule
                 \multirow{4}{*}{Air Quality} & NLL &  & & 1.49 &  & \underline{1.48} &  & \textbf{1.46} \\
                 & MQE & & & \underline{0.0235} & &  0.0312 &  & \textbf{0.0223}\\
                 & CRPS & & & \textbf{0.802} & &  0.832 &  & \underline{0.818}\\
                 \hdashline
                 & \textit{n. trees / depth} & & & \textit{100 / 5}  & & \textit{100 / 5} & & \textit{100 / 4}\\
                 \midrule
                 \multirow{4}{*}{Solar Flares} & NLL & & & 1.10 & & \textbf{-0.972} &  & \underline{-0.0234} \\
                 & MQE & & & \textbf{0.00185} & &  0.00597 &  & \underline{0.00367} \\
                 & CRPS & & & \underline{0.209} & &  0.212 &  & \textbf{0.195} \\
                 \cdashline{2-9}
                 & \textit{n. trees / depth} & & & \textit{25 / 7}  & & \textit{50 / 6} & & \textit{100 / 4}\\
                 \midrule
                 \multirow{4}{*}{Stock Performance} & NLL & & & \underline{-0.870} & &  -0.578 &  & \textbf{-1.04} \\
                 & MQE & & & \textbf{0.00124} & &  \underline{0.00324} &  & \textbf{0.00124} \\
                 & CRPS & & & \underline{0.0824} & &  0.0881 &  & \textbf{0.0777}\\
                 \cdashline{2-9}
                 & \textit{n. trees / depth} & & & \textit{50 / 4}  & & \textit{25 / 6} & & \textit{100 / 3}\\
                 \bottomrule
            \end{tabular}
            \label{tab:quality_metrics}
        \end{table}
        
        \textit{Efficiency} This work showed an algorithm modification that enables the DL8.5 algorithm to learn many trees at once, each optimizing a loss function with a different quantile parameter, while only exploring the tree space once. In this experiment, we measured, for different numbers of trees to learn, the execution time of QDL8.5 compared to the naive version that learns each tree by starting a new search every time. Figure \ref{fig:efficiencyplot} shows the results of this experiment. It shows that the naive version's execution time is linear with the number of trees. It also shows that the execution time of QDL8.5 is virtually independent of the number of trees. This shows that QDL8.5 can learn arbitrarily many trees, each describing a different point in the distribution range, at almost no additional computational cost compared to learning a single tree. We can also see that for 5 trees, the speedup of QDL8.5 is 4.74, which is close to the optimal speedup we could expect by learning the trees jointly. The same observation can be made for other numbers of trees. 

        \begin{figure}[!h]
        \centering
        \begin{minipage}{.35\textwidth}
            \includegraphics[width=0.95\textwidth]{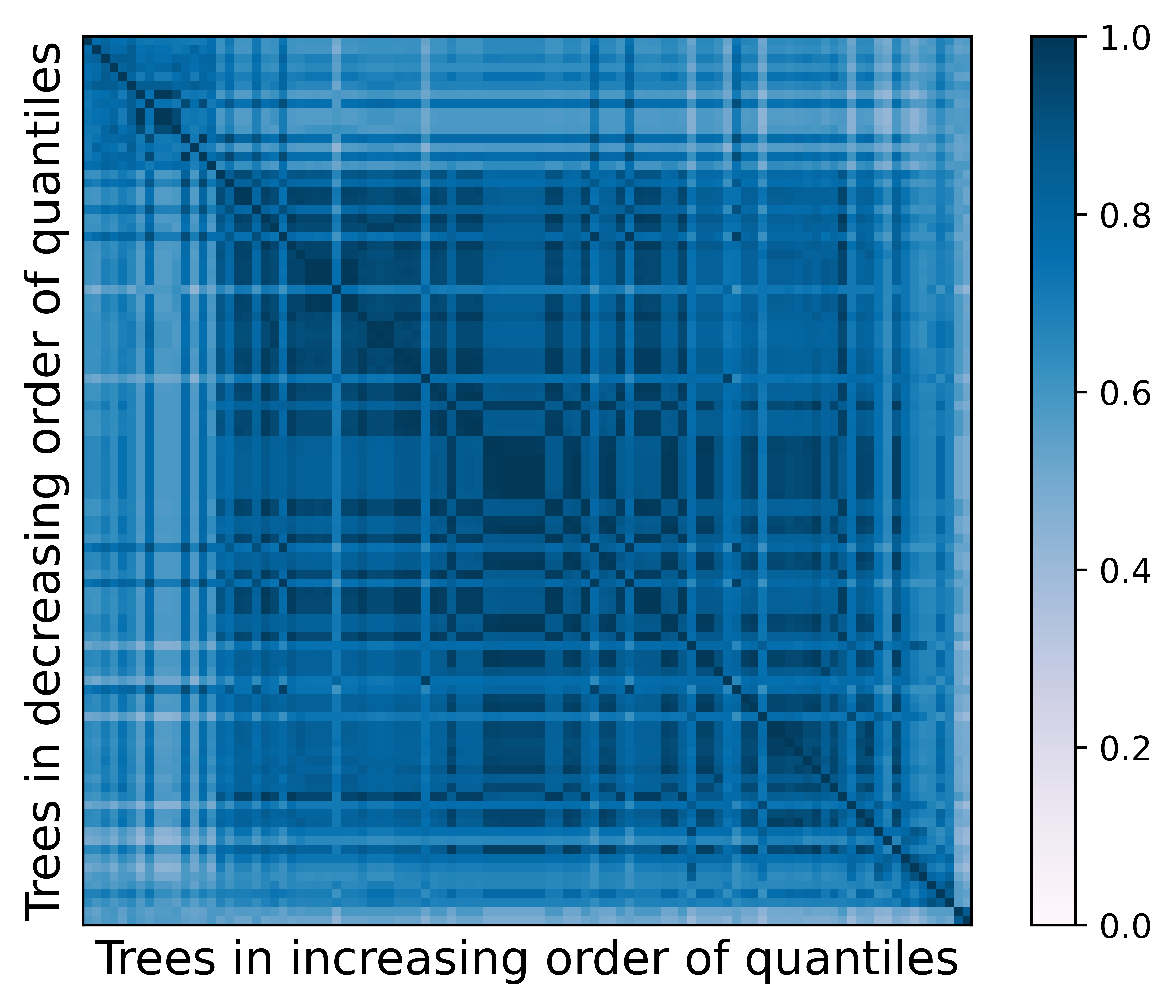}
            \caption{Similarity matrix of trees learned by QDL8.5 for the Air Quality dataset.}
            \label{fig:similarity matrix}
        \end{minipage}
        \quad
        \begin{minipage}{.55\textwidth}
            \includegraphics[width=0.95\textwidth]{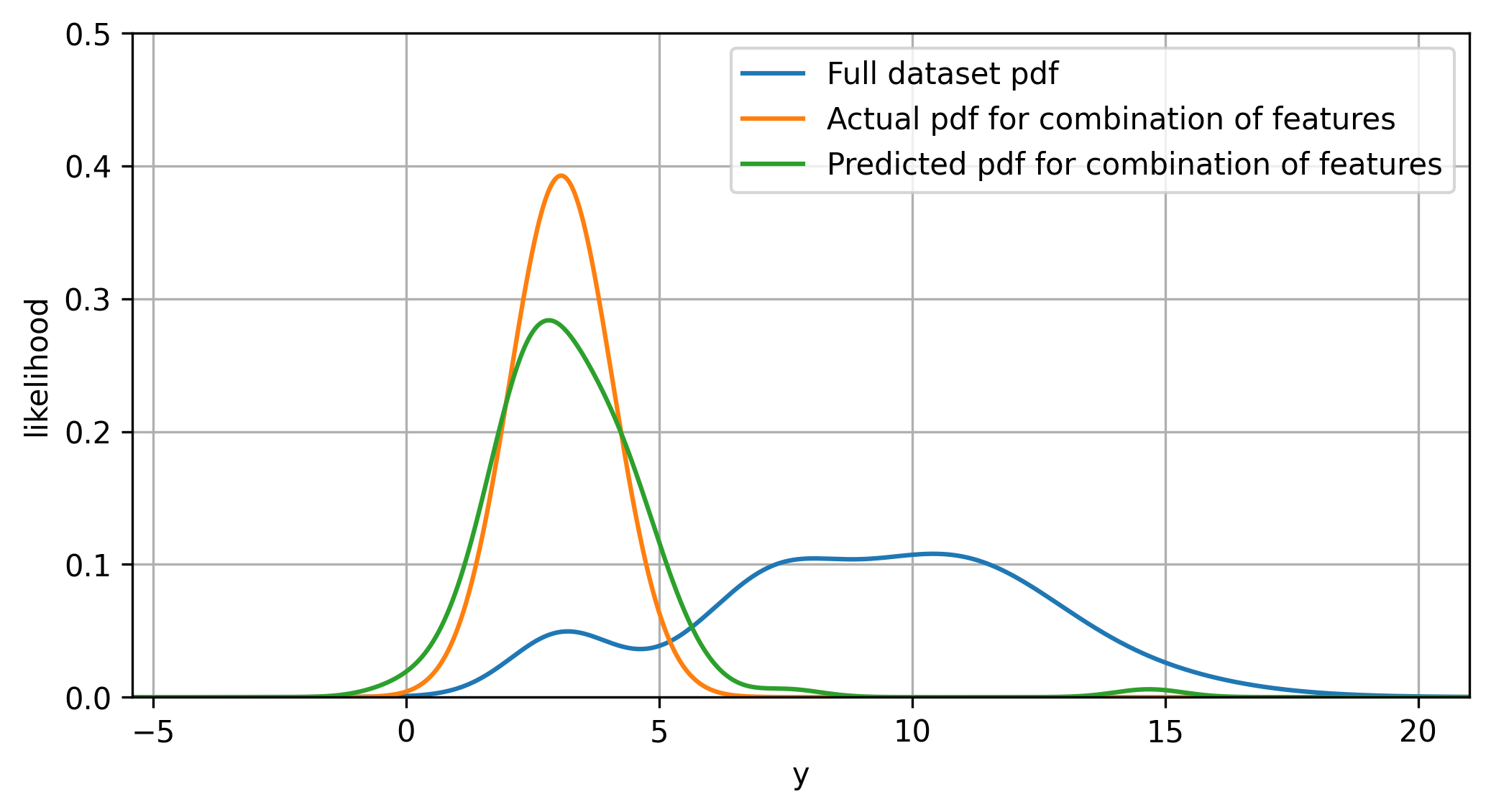}
            \caption{Distribution plots for a category within the synthetic dataset.}
            \label{fig:distribution plots}
        \end{minipage}
        \end{figure}

        \textit{Interpretability} Using optimal decisions is a way to provide inherently interpretable models. However, while we do generate shallow trees, we produce many of them, which may increase interpretability difficulty. However, if those trees happen to be similar, then inspecting only a few of them at different parts of the distribution range would reflect the majority of the information of the quantile regression ensemble. In this experiment, we have analyzed, for each tree, how it partitions the training dataset and performed a Jaccard index on those partitions to see how similar the trees are. The result of this experiment is shown in Figure \ref{fig:similarity matrix}. Based on this plot, a few observations can be made. First, generated trees are all quite similar, as the lowest Jaccard index in this matrix is $\sim 0.4$. Secondly, we can see that trees corresponding to close quantiles partition the feature space similarly, as attested by high Jaccard index values on blocks close to the diagonal. In this example, if we define \textit{tree zones} as being delimited by a change of 10\% in Jaccard index in successive trees, we would end up with 5 zones, and thus by picking only one among those zones, we can interpret \textit{whole} the distribution of the \textit{whole} data with only 5 trees of depth 4. 
        
        Finally, we can see that the tails of the distribution (low and high quantiles) are described by trees that differ from the centre of the distribution (large darker blocks in the centre and diagonal corners of the matrix), thus justifying the use of quantile regression to describe those parts of the distribution. Another way to interpret the results of the QDL8.5 algorithm is to plot the predicted conditional pdf. Figure \ref{fig:distribution plots} shows this for a particular category in the synthetic dataset. This figure shows that the predicted pdf is quite close to the actual pdf and allows a human to understand how the target variable behaves for a sample.

        Another point where the interpretability of our method is improved compared to other tree ensemble methods like Cadet RF and QRF is that in our case, each tree is linked with an interpretable value. Analyzing the tree corresponding to quantile 0.1 will give information about that specific part of the distribution for the whole dataset. For the other methods, all trees have to be analyzed to understand trends in the data, and their aggregation is not trivial. In addition, the optimality criteria ensures that the partitioning of the data is the best one for that quantile.

\section{Conclusion}
    This work presents a variation of the DL8.5 \cite{dl8.5} algorithm that enables it to perform simultaneous quantile regression, thus learning many trees at once, each describing a different part of the distribution while only exploring the search space once. This enables this model to generate shallow and interpretable decision trees that provide robust predictions via quantile regression and information about the complete conditional distribution of the data. Experiments have shown that this model achieves good quality matching or surpassing the state-of-the-art in this domain, that learning many trees comes at a small additional computing cost compared to learning only one, and finally, that the obtained trees are highly interpretable because they correspond to an interpretable parameter and because they only differ incrementally with the different quantiles.
%
%
%
\bibliographystyle{styling/splncs04}
\bibliography{bibliography}

\end{document}